%% file: main.tex
\documentclass{article} 
\usepackage{iclr2026_conference}

\input{math_commands.tex}

\usepackage{times}
\usepackage{graphicx}
\usepackage{hyperref}
\usepackage{url}
\usepackage{mathtools}
\usepackage{booktabs}
\usepackage{amssymb}
\usepackage{subcaption}

\newcommand{\para}[1]{\noindent\textbf{#1}}

\title{Why Can't Transformers Learn Multiplication? Reverse-Engineering Reveals Long-Range Dependency Pitfalls}

\makeatletter
\def\blfootnote{\xdef\@thefnmark{}\@footnotetext}
\makeatother

\author{
Xiaoyan Bai\thanks{Equal contribution. $^\dagger$Work done entirely at Harvard.}$^{\ \: 1}$,
Itamar Pres$^{*2}$,
Yuntian Deng$^3$,
Chenhao Tan$^1$,
Stuart Shieber$^4$,\\
\textbf{ Fernanda Viégas$^{4, 5\dagger}$,
Martin Wattenberg$^{4, 5\dagger}$,
Andrew Lee$^{4}$\blfootnote{testing}}\\
$^1$University of Chicago
$^2$MIT
$^3$University of Waterloo
$^4$Harvard University
$^5$Google DeepMind
}

\newcommand{\ATT}[2]{\ensuremath{\textsc{Att}^{#1}_{#2}}}
\newcommand{\MLP}[1]{\ensuremath{\textsc{MLP}^{#1}}}
\newcommand{\reals}{\mathbb{R}}

\arxiv

\begin{document}

\maketitle

\begin{abstract}
Language models are increasingly capable, yet still fail at a seemingly simple task of multi-digit multiplication.
In this work, we study why, by reverse-engineering a model that successfully learns multiplication via \emph{implicit chain-of-thought}, and report three findings:
(1) Evidence of long-range structure: Logit attributions and linear probes indicate that the model encodes the necessary long-range dependencies for multi-digit multiplication.
(2) Mechanism: the model encodes long-range dependencies using attention to construct a directed acyclic graph to ``cache'' and ``retrieve'' pairwise partial products.
(3) Geometry: the model implements partial products in attention heads by forming Minkowski sums between pairs of digits, and digits are represented using a Fourier basis, both of which are intuitive and efficient representations that the standard fine-tuning model lacks. 
With these insights, we revisit the learning dynamics of standard fine-tuning and find that the model converges to a local optimum that lacks the required long-range dependencies. 
We further validate this understanding by introducing an auxiliary loss that predicts the ``running sum'' via a linear regression probe, which provides an inductive bias that enables the model to successfully learn multi-digit multiplication.
In summary, by reverse-engineering the mechanisms of an implicit chain-of-thought model we uncover a pitfall for learning long-range dependencies in Transformers and provide an example of how the correct inductive bias can address this issue.
\ifarxiv
\blfootnote{Contact: \texttt{andrewlee@g.harvard.edu, smallyan@uchicago.edu}}
\blfootnote{Code: \url{https://github.com/ajyl/icot}}
\else
    \ificlrfinal
\blfootnote{Contact: \texttt{andrewlee@g.harvard.edu, smallyan@uchicago.edu}}
\blfootnote{Code: \url{https://github.com/ajyl/icot}}
    \fi
    \blfootnote{Code: \url{https://anonymous.4open.science/r/icot-F822}}
\fi
\end{abstract}

\input{1_introduction}
\input{2_experiment_setup}
\input{3_icot_mechanism}

\input{4_feature_geometry}

\input{5_debugging_sft}
\input{6_inductive_bias}
\input{7_related_work}

\input{8_conclusion}

\ifarxiv
  \input{8a_acknowledgements}
\else
  \ificlrfinal
    \input{8a_acknowledgements}
  \fi
\fi

\input{9_reproducibility}

\bibliography{main}
\bibliographystyle{iclr2026_conference}

\appendix
\input{appendix}

\end{document}

%% file: math_commands.tex
\usepackage{amsmath,amsfonts,bm}

\def\eqref#1{equation~\ref{#1}}

\def\1{\bm{1}}

\DeclareMathAlphabet{\mathsfit}{\encodingdefault}{\sfdefault}{m}{sl}
\SetMathAlphabet{\mathsfit}{bold}{\encodingdefault}{\sfdefault}{bx}{n}



%% file: 1_introduction.tex
\section{Introduction}
\label{sec:intro}

Large language models demonstrate striking capabilities across reasoning, planning, and tool use.
Yet, they also fail on surprisingly simple algorithmic tasks \citep{nye2021show, lee2023teaching}.
Why do Transformers excel at some tasks, but fail to learn others?
One such example is multi-digit multiplication.
Despite having \emph{billions} of parameters, models like Llama-3.2 90B or GPT4 still fail at 4x4-digit multiplication \citep{gambardella-etal-2024-language},\footnote{Note that some recent proprietary models that do solve multi-digit multiplication may rely on tool-use.} even when explicitly fine-tuned on the task \citep{yang2023gpt}.
Why do Transformers fail to learn multiplication?

We study these questions by contrasting a standard fine-tuned model (SFT), which fails at multiplication, with a model trained with \emph{implicit chain-of-thought} (ICoT) \citep{deng2024explicitcotimplicitcot, deng2023implicit}, which succeeds.
ICoT works by providing explicit chain-of-thought tokens during training, but gradually removes them and thus forces the model to internalize intermediate steps in its latent states.

We partially reverse-engineer the ICoT model and uncover several insights.
First, unlike the SFT model, the ICoT model learns the correct long-range structure needed for multi-digit multiplication. 
We provide evidence of this using logit attributions and linear regression probes.
\emph{Mechanistically}, the ICoT model encodes long-range dependencies by organizing its attention into a sparse, binary-tree-like graph, which (i) selects the correct digit pairs to compute partial products and (ii) ``caches'' these intermediate computations into earlier tokens for later retrieval.
Lastly, \emph{geometrically}, attention heads realize partial products as Minkowski sums of digit embeddings, and represent digits with Fourier bases, yielding a pentagonal prism structure -- both of which are intuitive and efficient representations that the SFT model lacks.

With these insights, we revisit the dynamics of standard fine-tuning: under gradient descent and an auto-regressive loss, the model never learns these long-range dependencies, and thus loss plateaus on the middle digits.
To confirm our understanding, we introduce a simple fix by introducing an auxiliary loss that supervises the model to predict a ``running partial sum'' through a lightweight linear regression probe.
This provides an inductive bias to learn the proper long-range dependencies, allowing it to achieve perfect accuracy, without any supervision from chain-of-thought.

In summary, by partially reverse-engineering a network that successfully implements multi-digit multiplication, we uncover how it implements long-range dependencies, a mechanism that the unsuccessful model lacks.
Our work highlights a challenge for Transformers to learn long-range dependency using gradient descent and an auto-regressive loss.
While we demonstrate a task-specific inductive bias to address this issue, we anticipate generic improvements to address this limitation.

%% file: 2_experiment_setup.tex
\section{Experiment Setup, Training ICoT, Notations}
\label{sec:setup}

\para{Task, Models.}
We are interested in understanding the difference in a model trained with standard fine-tuning and ICoT.
From experiments, we find that the simplest multi-digit multiplication in which standard fine-tuning fails but ICoT works is 4$\times$4 digit multiplications.
Similarly, the smallest architecture in which ICoT works is a 2-layer model with 4 attention heads.
Thus we carefully study a 2-layer 4-head ICoT model and a standard fine-tuned model trained on 4$\times$4 multiplication.

\para{Training Procedures.}
Our ICoT setup is the same as that \cite{deng2024explicitcotimplicitcot}.
Here we provide an informal overview of ICoT, with details in Appendix~\ref{appx_sec:icot_details}.
Namely, assume two operands $a =(a_3, a_2, a_1, a_0), b = (b_3,b_2,b_1,b_0)$ and their product $c = (c7\dots c_0)$.
Operands are written least-significant digit first, similar to other algorithmic setups~\citep{deng2024explicitcotimplicitcot, deng2023implicit, lee2023teaching}.

For ICoT, the training data includes intermediate chain-of-thought (CoT) tokens $q_i$ that explicitly record the step-by-step calculations.  
As a simple illustration, consider $12 \times 34$.  
The tokens appearing between the two equal signs follow the same CoT format used in our $4 \times 4$-digit multiplication tasks:

\vspace{-1.5em}
\begin{align*}
12 * 34 = \underbrace{48}_{12 *4} + \underbrace{360}_{12*30}\ \underbrace{(408)}_{\text{running sum}} = 408
\end{align*}

At each training epoch, a fixed number of CoT tokens are removed from the \emph{left} of the chain.  
Concretely, the training examples at each epoch may have the following form:
\begin{align*}
(\text{Epoch 1})&\quad a_0a_1a_2a_3 * b_0b_1b_2b_3\%\%\%\ q_0\ldots q_i \ldots q_j \ldots q_k\ldots q_{\tau}\ \#\#\#\#\ c_0\ldots c_7 \\
(\text{Epoch 2})&\quad a_0a_1a_2a_3 * b_0b_1b_2b_3\%\%\%\ q_i \ldots q_j \ldots q_k\ldots q_{\tau}\ \#\#\#\#\ c_0\ldots c_7 \\
(\text{Epoch 3})&\quad a_0a_1a_2a_3 * b_0b_1b_2b_3\%\%\%\ q_j \ldots q_k\ldots q_{\tau}\ \#\#\#\#\ c_0\ldots c_7 \\
\ldots \\
(\text{Epoch N})&\quad a_0a_1a_2a_3 * b_0b_1b_2b_3\%\%\%\ \#\#\#\#\ c_0\ldots c_7
\end{align*}
where $q_i$ are CoT tokens and $\%,\#$ are special delimiters.\footnote{These delimiters have no special meaning beyond matching the setup of \cite{deng2024explicitcotimplicitcot}.}
Note that after each epoch, the model sees a shorter chain by truncating some tokens, and that by the end, only the operands and final answer remain.
For comparison, standard fine-tuning only trains on the operands: $a_0a_1a_2a_3 * b_0b_1b_2b_3\%\%\% \#\#\#\#\ c_0\ldots c_7$.

Interestingly, the ICoT model is able to achieve 100\% accuracy on 4$\times$4 digit multiplication, while standard fine-tuning only achieves less than 1\% accuracy.
Note that scaling does not help -- scaling to a 12 layer 8 head model achieves the same $< 1\%$ accuracy, and \cite{yang2023gpt} show that fine-tuning a 2B model still plateaus at 95\% accuracy.

For more details regarding training (data format, sample size, hyperparameters), see Appendix~\ref{appx_sec:training_details}.

\paragraph{Notations.}
$\mathbf{h}^\ell_t$ indicates the hidden states at layer $\ell$ timestep $t$.
Timesteps for solution tokens $c_k, k=[0, \ldots, 7]$ are notated $t_{c_k}$.
\ATT{\ell}{h}$(\cdot)$, \MLP{\ell}$(\cdot)$ indicate the output of the attention heads or MLP blocks at layer $\ell$, head index $h$.
$E, U \in \reals^{V \times d}$ indicate (un)embedding weights.

%% file: 3_icot_mechanism.tex
\section{Comparing the Mechanisms of ICoT versus SFT}
\label{sec:mechanism}

\subsection{Long-range dependencies in multi-digit multiplication}
\label{subsec:key_components_in_multiplication}

\input{Figures/multiplication}
Here we discuss how one might solve multi-digit multiplication, and the required long-range dependencies needed to solve multiplication.

One approach to compute each digit, $c_k$, is as follows:

\vspace{-1.8em}
\begin{align}
    s_k \triangleq \underbrace{\sum_{i+j=k} a_ib_j,}_{\text{sum of partial products}} \quad
    c_k = (s_k + r_{k-1})\ \text{mod}\ 10, \quad
    r_{k} = \underbrace{\big\lfloor \frac{s_{k} + r_{k-1}}{10}\big\rfloor}_{\text{carry}}, \quad
    r_{-1} = 0
\end{align}

Note that both $c_k$ and $r_k$ can be expressed with an intermediary term $\hat{c}_k$, which encapsulates both the relevant information from the partial products and the carry:

\vspace{-1.2em}
\begin{align}
\label{eq:multiplication}
\hat c_k \triangleq s_k + r_{k-1},\qquad
c_k = \hat c_k \;(\mathrm{mod}\;10),\qquad
r_{k} = \big\lfloor \frac{\hat{c}_{k}}{10} \big\rfloor
\end{align}

Importantly, note the \emph{long-range dependencies} needed for multi-digit multiplication.
Specifically, we highlight two observations:
\textbf{(i)} To determine $c_k$, one must use all the partial products $\{a_ib_j | i+j\le k\}$, since all of these terms contribute to $c_k$.
\textbf{(ii)} Knowing the intermediary term $\hat{c}_k$ suffices to compute $c_k$ and to propagate necessary information for later digits.
Thus we use $\hat{c}_k$ as a probing signature (Section~\ref{subsec:long_range_dependency}) at each timestep $t_{c_k}$ to check if the model is utilizing all the necessary long-range information to predict the correct tokens $c_k$.

%
In the following sections, we demonstrate how the ICoT model satisfies such long-range dependency while the standard fine-tuning model does not.

\subsection{Evidence of Long-Range Dependencies in ICoT}
\label{subsec:long_range_dependency}

We first demonstrate two lines of evidence that the ICoT model satisfies long-range dependencies in multi-digit multiplication, while the standard fine-tuning model does not.

\para{Logit Attributions.}
Note from Figure~\ref{fig:multiplication} that digits $a_i, b_i$ can only affect $c_k$ terms where $k \geq i$.
Also note that at timestep $t_{c_k}$, the pairwise products $\{a_ib_j|i+j=k\}$ affect the final prediction $c_k$ the most.
``Earlier'' pairwise products $\{a_ib_j|i+j \leq k\}$ can still affect $c_k$, but with diminishing effects as $i+j$ gets smaller.

We directly test for these relationships in our ICoT and SFT models using logit attributions.
Namely, given an input sample $\textsc{orig} :=a_0a_1a_2a_3*b_0b_1b_2b_3$, we measure the logits of the model's predictions for $c_{0-7}: \text{logit}_{c_k}(\textsc{orig})$.
We then randomly swap out one of the operand digits at timestep $t$ (e.g., $\tilde{a}_2$) to construct a counterfactual input $\textsc{counter}_t =a_0a_1\tilde{a}_2a_3*b_0b_1b_2b_3$ and measure the change in logits: $\Delta_{t,k} = \text{logit}_{c_k}(\textsc{orig})-\text{logit}_{c_k}(\textsc{counter}_t)$
Thus $\Delta_{t,k}$ measures the effect that digit at timestep $t$ has on the prediction of token $c_k$.

\input{Figures/long_term_effects}

We use 1,000 samples for each $(t, k)$ pair and show the results in Figure~\ref{fig:long_term_effects}.
Note that for SFT, the model does not see the correct dependencies between earlier tokens to middle tokens, while the ICoT model does, suggesting that the model has indeed learned the correct long-range dependencies.

\para{Probing for $\mathbf{\hat{c}}_k$.}
Note from Figure~\ref{fig:multiplication} and Equation~\ref{eq:multiplication} that the long-range dependencies can be captured by an intermediate term, $\hat{c}_k$.
We test for whether $\hat{c}_k$ information can be decoded from the hidden states of the models using linear regression probes.
Namely, at each timestep $t_{c_k}$ we predict for $\hat{c}_{k}$ by training a single vector $\mathbf{w}_{k} \in \reals{}^{d}$ such that $\mathbf{w}_k\mathbf{h}_{t_{ck}}^{2.\text{mid}} = \hat{c}_{k}$ using a MSE loss, where $\mathbf{h}^{2.\text{mid}}$ is the hidden state at layer 2 after attention heads, before MLPs.

Figure~\ref{fig:c_hat_probe} reports the mean absolute error from probing for $\hat{c}_k$ for middle and late digits, $k=2,\ldots,6$.
Note that the accuracy from the ICoT model is much higher than that of SFT, further suggesting that the ICoT model has learned the correct long-range dependencies while SFT has not.

\input{Figures/c_hat_probe}

\subsection{Encoding Long-Range Dependencies via Attention Trees}
\label{subsec:binary_trees_via_attention}

How does the ICoT model compute long-range dependencies?
Here we describe how the model's attention patterns induce a shallow directed acyclic graph, akin to a binary expression tree, in order to encode long-range dependencies.

Namely, in the first layer, across all timesteps $t > 5$,\footnote{Note that only after timestep 5, both $a$ and $b$ tokens appear in the context.} each attention head only attends to a \emph{pair} of digit tokens, $\{a_i, b_j\}$ (Figure~\ref{fig:attn_tree}, left).
This allows the model to produce the pairwise product $a_ib_j$ (see Section~\ref{subsec:minkowski_sums} for \emph{how} attention heads represent pairwise products), but also allows the model to \emph{cache} the product $a_ib_j$ in the hidden state of layer 1 at timestep $t$ (i.e., $\mathbf{h}^1_t$).
Put differently, product pairs $\{a_ib_j\}_{i,j\in\{0, \ldots 4\}}$ are ``cached'' in the first layer across different timesteps ($\mathbf{h}^1_t, t < t_{c_k}$).

At later timesteps $t \geq t_{c_k}$, when the model predicts solution tokens $c_k$, this allows the second layer attention heads to attend to a small set of previous \emph{cache sites}, i.e., where the appropriate pairs of products $a_ib_j, i+j=k$ are stored from earlier timesteps.

\input{Figures/attention_tree}

\para{Example:}
Figure~\ref{fig:attn_tree} depicts the attention patterns when the model predicts $c_2$, given input ``$a_{0\ldots 3}*b_{0\ldots 3} = c_0c_1$''.
These attention maps are averaged from 1,000 samples from a held out test set.
The necessary terms to compute $c_2$ are $a_2b_0, a_1b_1, a_0b_2$, and $\hat{c}_1$ (which in turn requires $a_1b_0, a_0b_1, a_0b_0$).

Attention heads \ATT{2}{3}, \ATT{2}{4} each attend to positions $(b_0, b_2, c_1)$ and $(b_3, \text{``}\#\text{''}, c_0)$.
Inspecting what was ``cached'' in the first layer at those timesteps reveals the necessary partial products to compute $c_2$.
For example, at timestep $b_0$, \ATT{1}{1}, \ATT{1}{2} attend to $a_2, b_0$; at timestep $b_2$ \ATT{1}{1} attends to $a_1, b_1$ while \ATT{1}{2} attends to $a_0, b_2$; at timestep $c_0$ \ATT{1}{1} attends to $a_1, b_0$, \ATT{1}{2} attends to $a_0b_1$.
Thus the model can derive partial products, $a_2b_0, a_1b_1, a_0b_2, a_1b_0, a_0b_1$ with its attention tree.\footnote{Note that there may be a couple of different ways that $a_0b_0$ is derived. One possibility is to re-use $a_0, b_0$ information that was fetched at various timesteps. Another possibility is when $a_0$ is slightly attended to at \ATT{2}{3} (difficult to see in our visuals). Note that $a_0b_0$ plays a relatively minor role in computing $c_2$ compared to all other partial products.}

While Figure~\ref{fig:attn_tree} shows an example of the ``attention tree'' for predicting $c_2$, one can similarly reconstruct the correct trees for all digits $c_0, \ldots, c_7$ using the attention patterns for all digits in Figure~\ref{fig:attn_tree_app}.

In summary, for each output step $c_k$, the ICoT model constructs a binary-tree-like graph, spread out across timesteps, to attend to the correct pairs of tokens, allowing it to compute partial products.

%% file: Figures/multiplication.tex
\begin{figure}[t]
\vspace{-3em}
\begin{center}
\includegraphics[width=0.8\textwidth, trim={9cm, 4.5cm, 7.5cm, 5.5cm},clip]{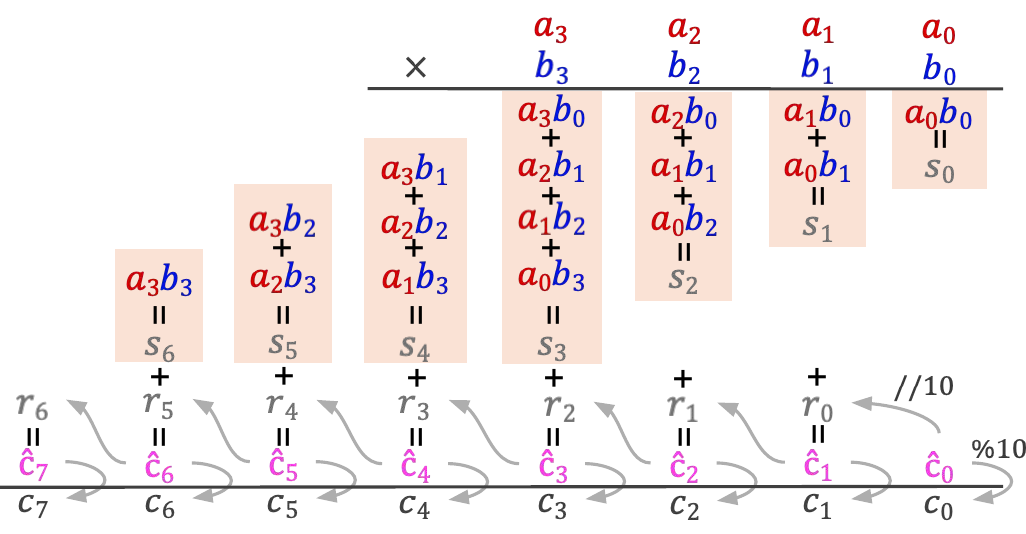}
\end{center}
\caption{\label{fig:multiplication}
\textbf{Multiplication has long-range dependencies}, which can be captured by an intermediate value $\hat{c}_i$, from which both the solution ($c_i$) and carries ($r_i$) can be derived from.
\vspace{-10pt}
}
\end{figure}

%% file: Figures/long_term_effects.tex
\begin{figure}
\begin{center}
\includegraphics[width=0.95\textwidth]{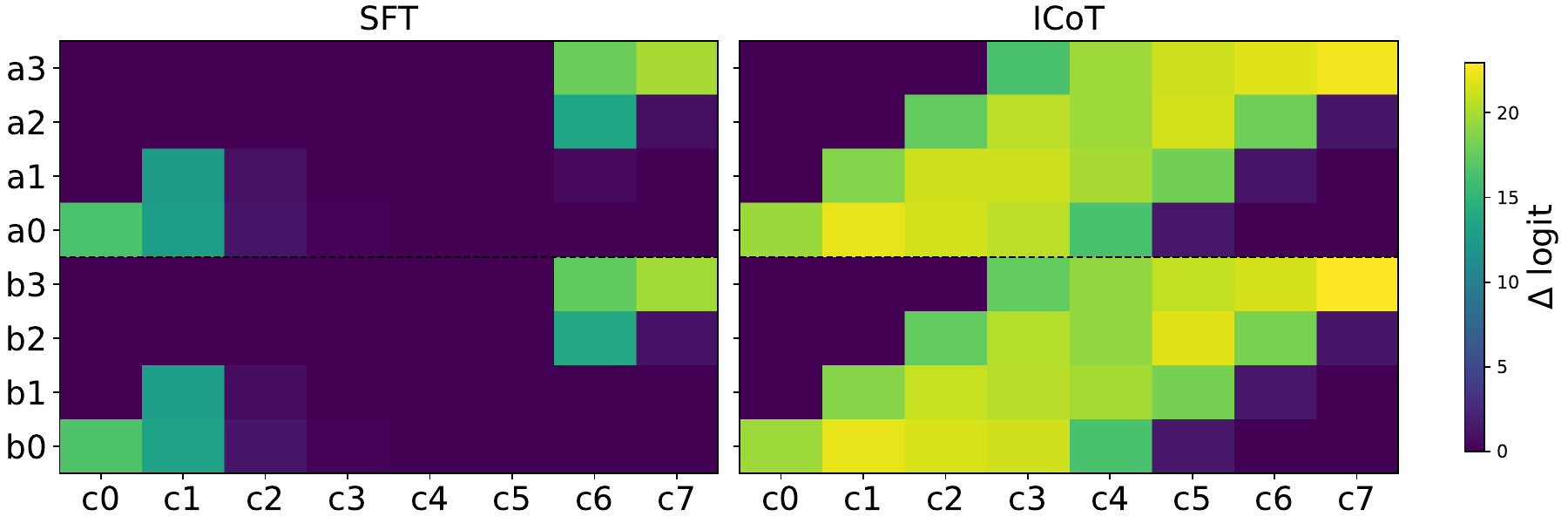}
\end{center}
\vspace{-10pt}
\caption{\label{fig:long_term_effects}\textbf{Logit Attribution.}
We test for whether each model has correctly learned long-range dependencies by measuring how sensitive the logits of output digits $c_i$ are to each operand digit (i.e., $a_i, b_j$).
This is done by measuring the change in $c_i$'s logits when a single operand digit is perturbed.
\vspace{-2.2em}
}
\end{figure}

%% file: Figures/c_hat_probe.tex
\begin{figure}
\begin{center}
\includegraphics[width=0.99\textwidth]{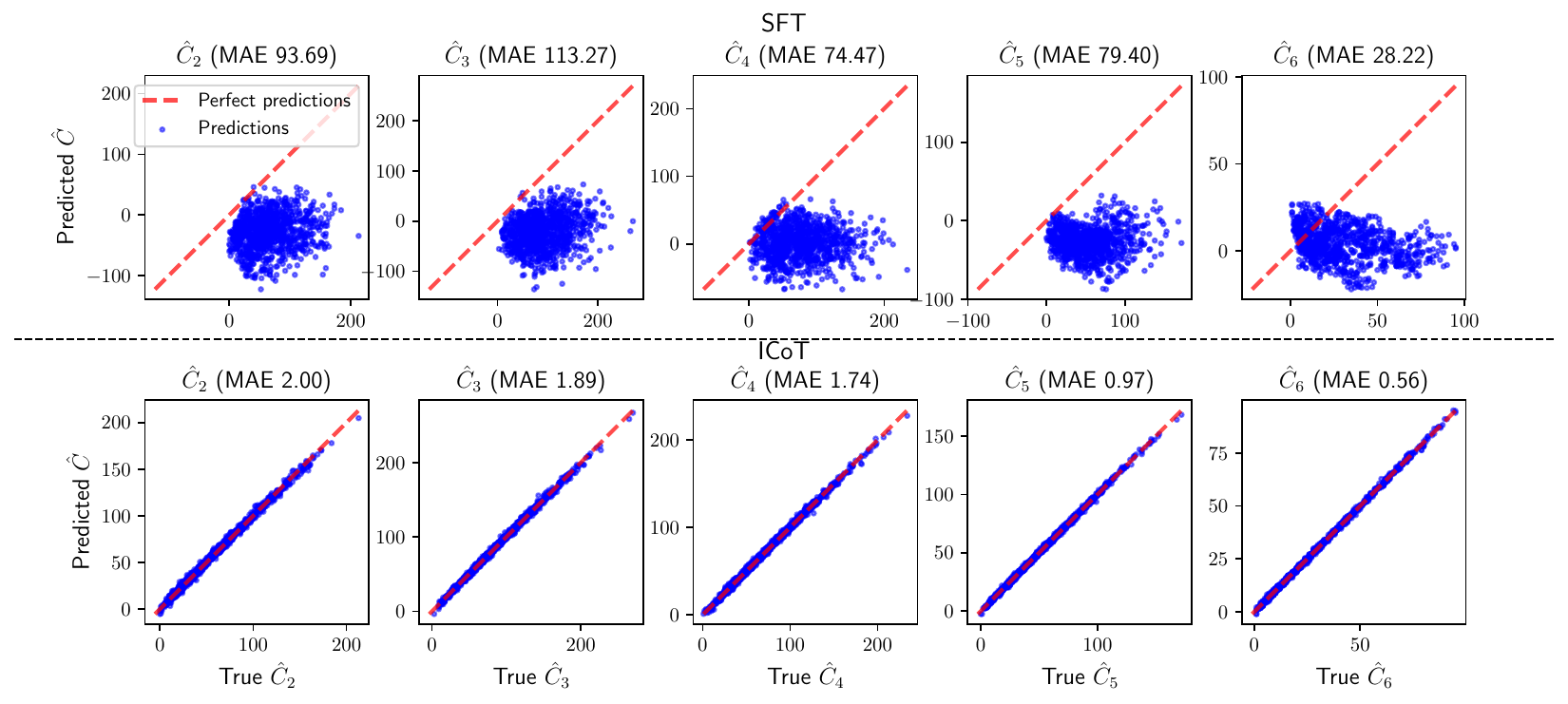}
\end{center}
\vspace{-10pt}
\caption{\label{fig:c_hat_probe}\textbf{Linear regression probing results for $\hat{c}$.}
We probe from the middle of the last Transformer block, after attention heads but before MLPs.
\vspace{-1em}
}
\end{figure}

%% file: Figures/attention_tree.tex
\begin{figure}
\begin{center}
\includegraphics[trim={3.5cm 2cm 3.5cm 2cm}, width=\textwidth]{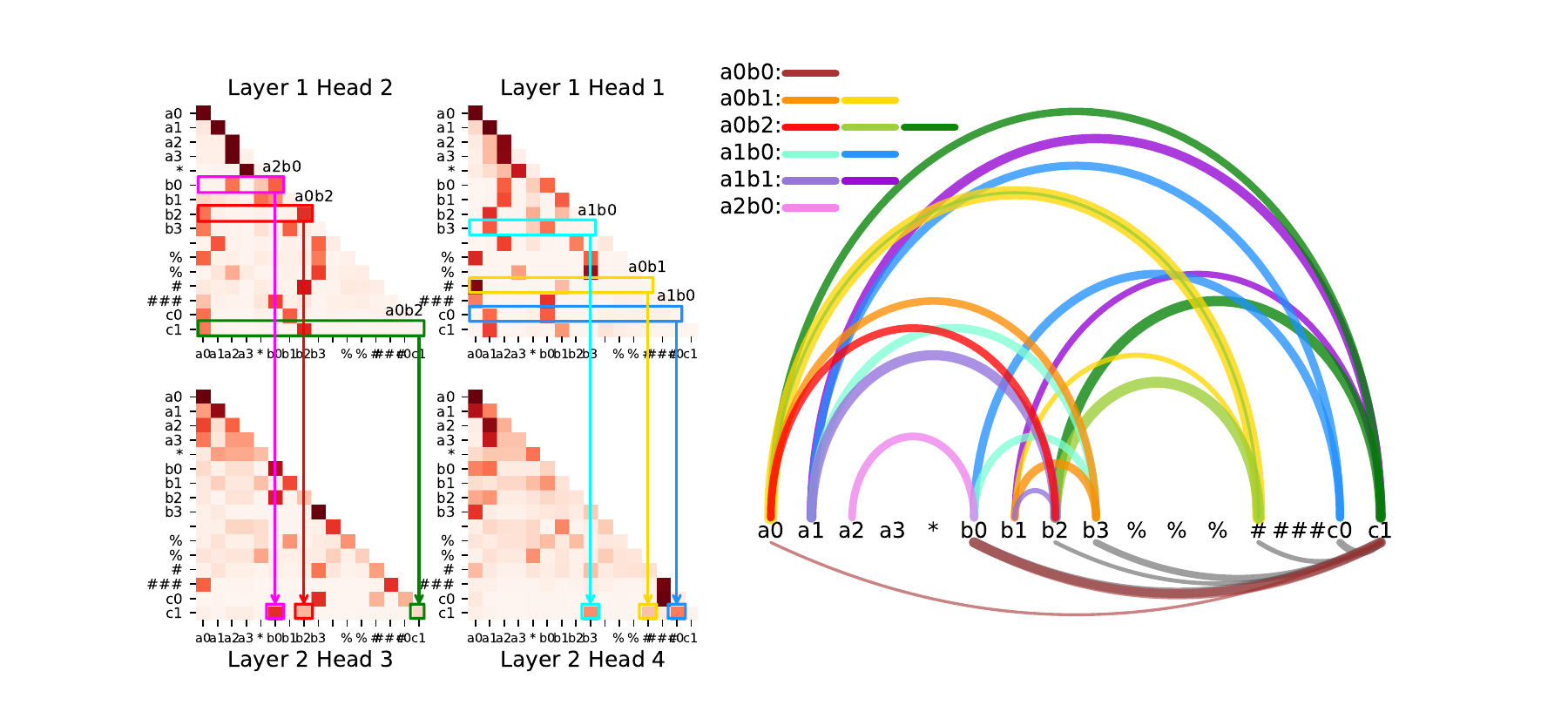}
\end{center}
\vspace{1.2em}
\caption{\label{fig:attn_tree}\textbf{Visualization of attention tree to compute $\textbf{c}_2$.}
Left: Attention maps for selected heads show the first layer ``cache'' pairwise products ($a_ib_j$) across earlier timesteps, from which the second layer reads from (Not all tree paths are shown).
Right: A visualization of the attention tree. Each arc indicates tokens being attended to at specific timesteps.
Colored arcs above and below the digits indicate attention patterns from the first and second layers respectively.
Example: orange arc indicates that at timestep $b_3$, the model attends to $a_0$ and $b_1$, from which the second layer reads from.
\vspace{-2em}
}
\end{figure}

%% file: 4_feature_geometry.tex
\section{Feature Geometry of ICoT}
\label{sec:feature_geometry}

In addition to the mechanisms seen in Section~\ref{sec:mechanism}, we also study the \emph{geometry} of features in ICoT.

\subsection{Digit-wise Multiplications as Minkowski Sums}
\label{subsec:minkowski_sums}

Note from Section~\ref{subsec:binary_trees_via_attention} that the attention patterns are sparse, often only attending to the two digits $a_i, b_j$ being multiplied.
In such a case, the outputs of the attention head form a Minkowski sum.

Namely, consider a single head \ATT{1}{}($i, j$) at the first layer, attending to two digits $a_i$, $b_j$.
Let $W_O \in \reals{}^{d \times d_{head}}, W_V \in \reals{}^{d_{head}\times d}$ be the output and value weights of the attention head, $E[a_i] \in \reals{}^d$ the token embedding for token $a_i$, and $A_i := W_OW_VE[a_i], B_j:=W_OW_VE[b_j], A_i, B_j \in \reals{}^d$.

In such a case, when the model spends $\alpha\%$ of its attention on digit $a_i$, and thus attends to digit $b_j$ by $(1-\alpha)\%$, the set of all possible values for the output of the attention head forms a Minkowski sum:

\vspace{-2em}
\begin{align}
    \ATT{1}{}(i, j) &= \alpha A_i + (1-\alpha)B_j + \epsilon
    \\
    \{\ATT{1}{}(i, j)\}_{i, j} &\subseteq (\alpha A) \oplus ((1-\alpha)B) \oplus \epsilon 
\end{align}
(ignoring position embeddings).
See Figure~\ref{fig:attn_3d_pcas} (a) for a visualization.
 
\input{Figures/attn_3d_pcas}

Visually, 3D PCAs can reveal nested representations.
Namely, we can observe clusters, each cluster corresponding to a feature (i.e., $a_i$).
These clusters form a ``global'' geometry.
When zoomed in to each cluster, we observe additional clusters for a second feature (i.e., $b_j$) that form a ``local'' geometry of the same shape as its global counterpart.
See Figure~\ref{fig:attn_3d_pcas} (b-d) for examples.

This observation can be explained by deconstructing the covariance of the attention output:

\vspace{-1.5em}
\begin{align}
    \Sigma_{\ATT{}{}} = \alpha^2\Sigma_A + (1-\alpha)^2\Sigma_B,
\end{align}
where $\Sigma_A = \text{Cov}(A_i), \Sigma_B = \text{Cov}(B_j)$.
First, note that if we ignore positional encodings, $\Sigma_A$ and $\Sigma_B$ share the same eigenvectors, as they each depend on the same terms ($E[\cdot], W_O, W_V$), which are picked by PCA.
Further note that fixing a value for $a_i$ leaves a local covariance, $\Sigma_{local|a_i} = (1-\alpha)^2\Sigma_B$, which again share the same eigenvectors with the global $\Sigma_{\ATT{}{}}$ term, leading to the same local geometry when projected onto.

\subsection{Embedding Digits on a Pentagonal Prism via Fourier Bases}
\label{subsec:fourier_bases}

Similar to \cite{kantamneni2025language}, we find that our model encodes digits in Fourier space. 
Specifically, the model’s embeddings $E$, the final hidden layer $\mathbf{h}^L$, and even the weights of the last MLP can be well reconstructed from a small set of Fourier basis functions.

\input{Figures/fourir_basis}

Figure~\ref{fig:fourier_basis} shows a 3D PCA visualization of the final hidden layer at timestep $t_{c_2}$, for both the SFT and ICoT models.
While the SFT hidden states do not reveal any obvious patterns, the ICoT hidden states reveal a striking pattern: the ten digits form vertices of a \emph{pentagonal prism}.

This structure is naturally explained by Fourier modes. 
Consider the Fourier expansion
\[
\sum C_n*e^{-2\pi i \tfrac{kn}{10}}, \quad n=0,\ldots,9.
\]

where $C_n (\neq c_k)$ is some constant per digit $n$.
Following \cite{kantamneni2025language}, we take frequencies $k \in \{0,1,2,5\}$, yielding the real Fourier basis
\[
\Phi(n) =
\left[
\begin{array}{c@{\quad}c@{\quad}c@{\quad}c@{\quad}c@{\quad}c}
\mathbf{1}(n)
& \cos\!\left(2\pi\tfrac{n}{10}\right) & \sin\!\left(2\pi\tfrac{n}{10}\right)
& \cos\!\left(2\pi\tfrac{n}{5}\right)  & \sin\!\left(2\pi\tfrac{n}{5}\right)
& \boldsymbol{p}(n) 
\\[-2pt]
\scriptstyle (k=0) & \scriptstyle (k=1) & \scriptstyle (k=1) &
\scriptstyle (k=2) & \scriptstyle (k=2) & \scriptstyle (k=5)
\end{array}
\right],
\]
where $\mathbf{1}(n)\equiv 1$ (the DC component) and $\boldsymbol{p}(n)\equiv (-1)^n$ (the Nyquist/parity vector). 
The sine terms for $k=0$ and $k=5$ vanish over $n=0,\dots,9$ and are omitted.

The final hidden layer $\mathbf{h}^L$ can be reconstructed via these six terms (see Appendix~\ref{appx_sec:fourier-structure}), indicating that the final hidden state is encoded using Fourier bases.

Revisiting Figure~\ref{fig:fourier_basis}, the first principal component (PC1) aligns with the parity vector $\boldsymbol{p}(n)$, separating even from odd digits. 
Second and third principal components span the $k=2$ Fourier pair ($\cos, \sin(\tfrac{2\pi n}{5})$), so the digits lie on two regular pentagons: one each for even and odd digits. 
The digits within each pentagon advance by $n+4\pmod{10}$ (e.g., $n = 0\!\to\!4\!\to\!8\ldots$, same for odd digits), allowing a walk around the pentagon while staying within the even/odd set.
Interestingly, taking $\pmod 5$ on such a sequence yields decreasing steps of 1 ($n \pmod 5 = 0\!\to\!4\!\to\!3\ldots$).
Lastly, the two pentagons are parallel and stacked along PC1, with corresponding vertices differing by \(\pm 5\) (same phase, opposite parity). 
Together, these yield the pentagonal-prism geometry in Figure~\ref{fig:fourier_basis}.

%% file: Figures/attn_3d_pcas.tex
\begin{figure}
\begin{center}
\includegraphics[width=0.99\textwidth]{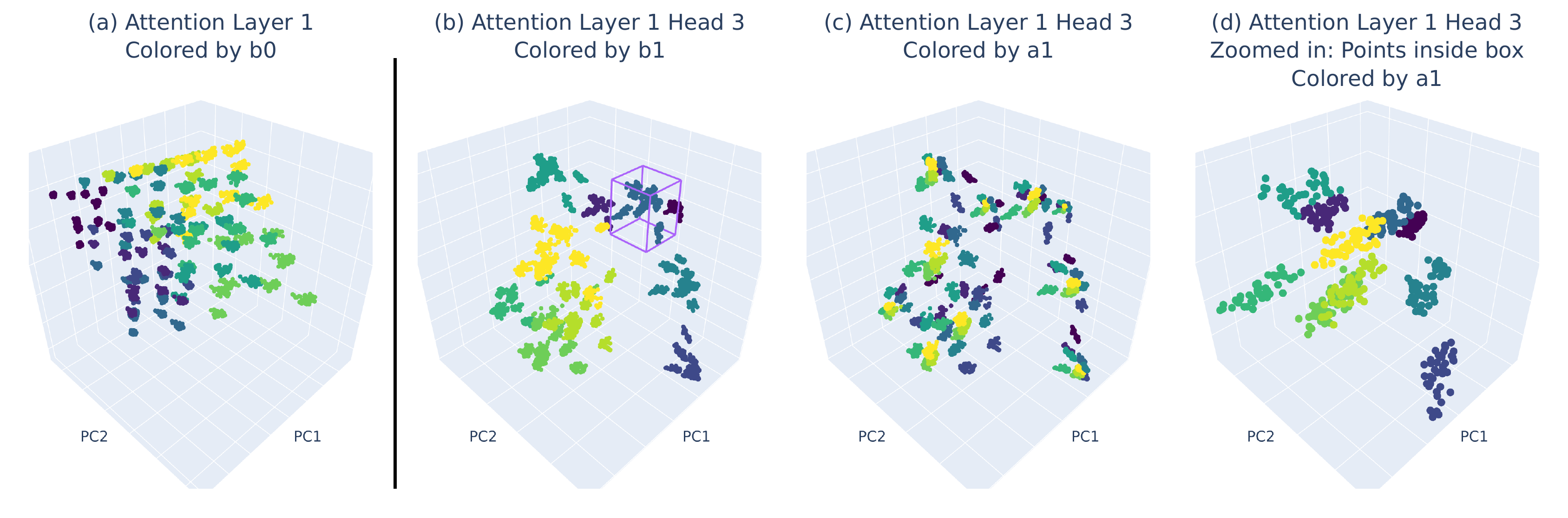}
\end{center}
\vspace{-10pt}
\caption{\label{fig:attn_3d_pcas}\textbf{3D PCA of attention head outputs can form Minkowski sums}, which in turn can form nested representations.
Each color represents a different digit.
\vspace{-10pt}
}
\end{figure}

%% file: Figures/fourir_basis.tex
\begin{figure}
\begin{center}
\includegraphics[width=0.99\textwidth]{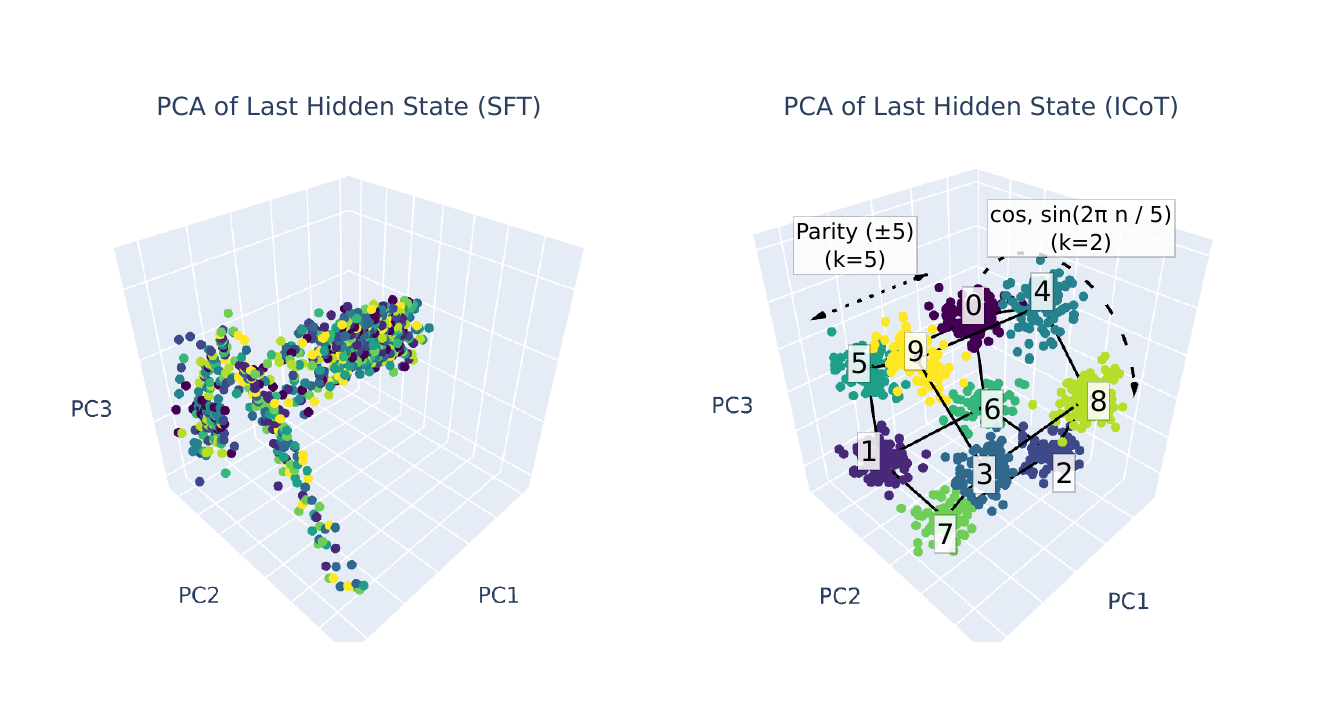}
\end{center}
\vspace{-2em}
\caption{\label{fig:fourier_basis}\textbf{Digits embedded in a pentagonal prism, using Fourier bases.}
No obvious patterns in the SFT model, but the ICoT model encodes digits in a pentagonal prism using Fourier bases.
\vspace{-10pt}
}
\end{figure}

%% file: 5_debugging_sft.tex
\section{Pitfalls of Learning: Lack of Long-Range Dependency}
\label{sec:debugging_sft}

Given our insights, we revisit why Transformers fail at multiplication under standard fine-tuning.

In particular, in Figure~\ref{fig:grad_norms_and_losses} (a), we inspect the gradient norms (top row) and losses (bottom row) \emph{per token} $c_k$ over the course of training.
There are a few observations to make.

\input{Figures/grad_norms_and_losses}

First, note from the loss curves that the first two digits, $c_0, c_1$, followed by the last digit, $c_7$, are learned first, as indicated by their immediate drop in loss to near zero.
This aligns with the gradient norms observed for these tokens: within the first few steps, these tokens receive gradients, but their norms quickly drop to near zero once the loss for these tokens reach zero.
Also note that the order in which tokens are learned according to gradient norms and losses is consistent.

The model then eventually learns to predict $c_2$.
However, middle digits, $c_3$ to $c_6$ are never learned.
Despite only the middle digits receiving gradients (as they are the only sources of loss remaining), their losses plateau, suggesting that the model is stuck in a local optimum that lacks the long-range dependencies to properly learn the middle digits.

Note that scaling to a larger model does not address this issue, as the same pattern can be found in a 12 layer 8 head model: see Appendix~\ref{appx_sec:grads_and_norms_12L}.

%% file: Figures/grad_norms_and_losses.tex
\begin{figure}
\begin{center}
\includegraphics[width=0.99\textwidth]{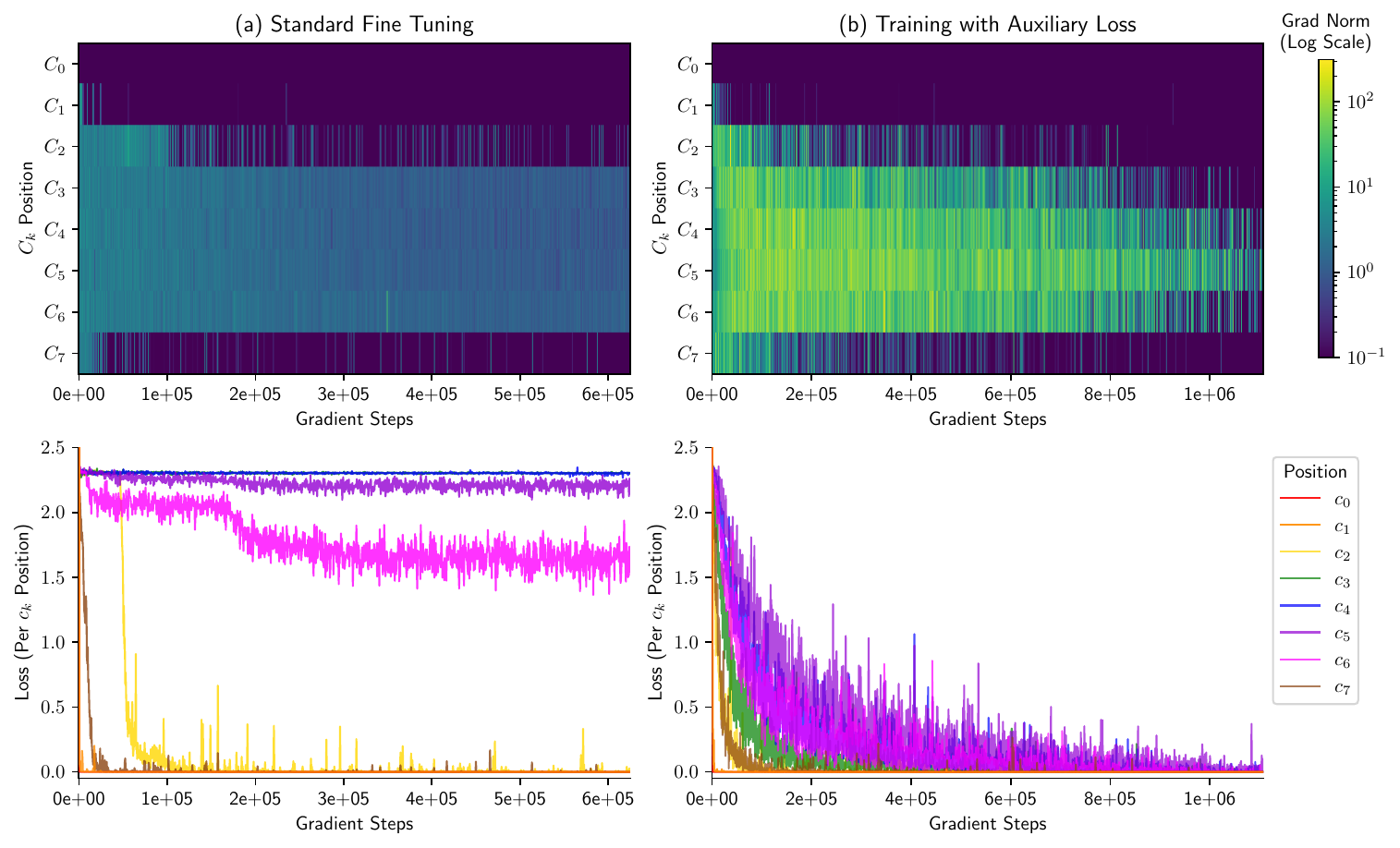}
\end{center}
\vspace{-10pt}
\caption{\label{fig:grad_norms_and_losses}\textbf{
Gradient norms and losses \emph{per token} $c_k$.}
While both methods learn digits $c_0, c_1, c_7$ first, standard fine-tuning gets stuck in a local optimum without having learned the right long-range dependencies, while training with the auxiliary loss allows the model to learn the middle digits.
\vspace{-10pt}
}
\end{figure}

%% file: 6_inductive_bias.tex
\section{Learning Multiplication Without ICoT}
\label{sec:inductive_bias}

To further validate our understanding of why Transformers fail to learn multiplication, we demonstrate an example of a simple fix to teach Transformers multiplications without needing ICoT.

In particular, we leverage the observation from Section~\ref{subsec:key_components_in_multiplication} in that ($i$) multi-digit multiplication requires long-range dependencies between digit $c_k$ and pairwise products $\{a_ib_j|i+j \leq k\}$, and ($ii$) such dependency can be summarized by an intermediate value $\hat{c}_k$ to produce $c_k$.

Thus in order to guide the Transformer to learn long-range dependencies, we simply add an auxiliary loss term to predict $\hat{c}_k$ at each timestep $t_{c_k}$.
We attach an additional linear regression head $\mathbf{w}_h \in \reals^{d}$ to the output of $H (=2)$ attention heads in the second layer.
These regression heads are trained to predict the correct accumulated sum $\hat{c}_{k}$ at each timestep $t_{c_k}, c_k\in[0, \ldots 7]$ with a MSE loss:
\begin{align}
    z_{i}^h &= \mathbf{w}^\top_h\ATT{2}{h}(\cdot)\\
    \mathcal{L}_{aux} &= \frac{1}{H}\sum_{h \in H}\frac{1}{8}\sum_{i=0}^7(z_i^h - \hat{c}_i)^2 \\
    \mathcal{L} &= \mathcal{L}_{LM} + \lambda\mathcal{L}_{aux}
\end{align}

where $\mathcal{L}_{LM}$ is the standard language modeling loss.

This introduces an inductive bias for the task, and allows our 2-layer model to correctly learn 4x4 multiplication with 99\% accuracy.
Again, note that a larger 12-layer model still fails at multiplication under standard fine-tuning.

Revisiting Figure~\ref{fig:grad_norms_and_losses} (b) demonstrates a very different learning dynamic.
We observe the model learn early and last digits ($c_0, c_1, c_7$) and work inwards ($c_2, c_3, c_4, c_6$, and finally $c_5$).

\para{Limitation.}
Obviously the suggested inductive bias pertains specifically to our task.
However, our experiments demonstrate the pitfall of Transformers that require long-range dependencies, and that it is possible to overcome such a pitfall with the correct inductive biases.
We speculate that there are other generalizing inductive biases that can improve performance on tasks with long-range dependencies \citep{tay2020long}, and leave this for future work.

\input{Figures/inductive_bias_attn}

\subsection{Does the model with auxiliary loss learn the same mechanism as ICoT?}
\label{subsec:inductive_bias_mechanism}

A natural question that arises is whether ICoT and our inductive bias leads to the same mechanisms.

Inspecting the attention patterns suggests that a similar (but not necessarily exact) mechanism is learned: see Figure~\ref{fig:inductive_bias_attn}.
Namely, the model similarly forms an ``attention tree'' to sparsely attend to the correct pairs of digits for each $c_i$ in the first layer (red boxes).
Interestingly, in the auxiliary-loss model we also observe an attention head (Layer 2 Head 2) that simultaneously attends to all the necessary digits, $\{a_{i\leq k}, b_{i\leq k}\}$, at each timestep $t_{c_k}$, forming a parallelogram-like attention pattern (black box) akin to the shape seen in Figure~\ref{fig:long_term_effects}.

%% file: Figures/inductive_bias_attn.tex
\begin{figure}
\begin{center}
\includegraphics[width=0.99\textwidth]{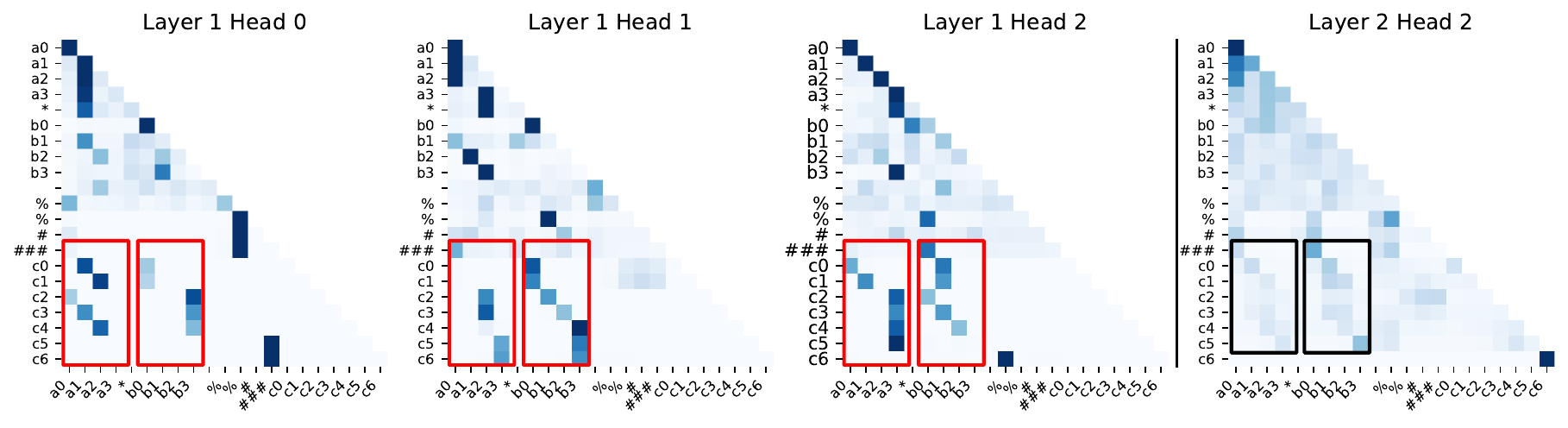}
\end{center}
\vspace{-10pt}
\caption{\label{fig:inductive_bias_attn}\textbf{Attention pattern of model trained with auxiliary loss.}
This model similarly produces a ``binary attention tree'' (red boxes), but interestingly, we also see an attention head that attends to all necessary pairwise digits simultaneously (black box), producing a pattern akin to Figure~\ref{fig:long_term_effects}.
\vspace{-5pt}
}
\end{figure}

%% file: 7_related_work.tex
\section{Related Work}

\paragraph{Studying Transformers with Arithmetic Tasks.}
A growing line of work study Transformers under controlled settings to better characterize their behavior~\citep{allen2023physics, allen2023physics3, li2023emergent, nanda2023emergent, park2024competition, park2024iclr}.
Often, arithmetics is a natural and popular domain~\citep{lee2023teaching, ye2024physics, nikankin2024arithmetic}, which has led to numerous insights.
For instance, \cite{nanda2023progress} study how Transformers perform modular addition to explain grokking.
\cite{kantamneni2025language} find that large language models use trigonometry to do addition, encoding digits using Fourier bases, while \cite{nikankin2024arithmetic} suggest that they also rely on heuristics.
\cite{cai2025extrapolation} study length generalization in Transformers using arithmetic tasks.
Similarly, we study the limitations of Transformers by studying why it fails to learn multi-digit multiplication.

\para{Process Supervision.}
Recent work trains models with \emph{process supervision}, in which feedback is given not just on final correctness but on each intermediate reasoning step.  
For example, \citet{uesato2022process} demonstrate that process-supervision can yield less reasoning errors on GSM8K compared to outcome-only supervision.
Similarly, \citet{lightman2023letsverify} show that step-level human feedback on MATH leads to stronger reward models.
More recently, \citet{zhong2023mathshepherd}'s Math-Shepherd automates step-wise rewards via continuation-based verification, improving performance on both GSM8K and MATH.
ICoT similarly plays the role of process supervision in latent space, by slowly removing chain-of-thought tokens during training such that the model internalizes the reasoning procedure.
We thus use ICoT's success on multiplication to study why Transformers fail.

%% file: 8_conclusion.tex
\section{Conclusion}
In this work, we study why Transformers fail on a seemingly simple task of multi-digit multiplication.
We answer this question by reverse-engineering a model trained with implicit chain-of-thought, and uncover that it has learned to compute the correct long-range dependencies needed for multi-digit multiplication. 
Our findings point to a pitfall of the standard recipe for training language models: using gradient descent with an auto-regressive loss on Transformers does not encourage the model to learn the right long-range dependencies.
While we provide a simple example of how the right inductive bias can address such a limitation, we anticipate future work to provide a generic solution to improve on tasks with long-range dependencies.

%% file: 8a_acknowledgements.tex
\subsubsection*{Acknowledgments}
XB, CT acknowledge the support of NSF grants IIS-2126602.
YD is supported by an NSERC Discovery Grant (RGPIN-2024-05178) and a Starter Grant from the University of Waterloo.
AL, MW, and FV acknowledge support from the Superalignment Fast Grant from OpenAI, Effective Ventures Foundation, Effektiv Spenden Schweiz, and the Open Philanthropy Project.
IP thanks Keya Hu for constructive feedback and AL thanks Thomas Fel for fruitful discussions regarding feature geometry.

%% file: 9_reproducibility.tex
\subsection*{Reproducibility Statement}

Our code to reproduce all of our experiments can be found in 
\ifarxiv
\url{https://github.com/ajyl/icot}
\else
\ificlrfinal
\url{https://github.com/ajyl/icot}
\fi
\url{https://anonymous.4open.science/r/icot-F822/}.
\fi.
Appendix~\ref{appx_sec:training_details} provides details of our training setup, including data formats, sample size, and hyperparameters.

%% file: appendix.tex
\input{appendix/icot_training}
\input{appendix/fourier}

\input{appendix/grads_and_losses_12L}
\input{appendix/attn_tree_dependency}
\input{appendix/llm_usage}

%% file: appendix/icot_training.tex
\section{Training Details}
\label{appx_sec:training_details}

Here we provide additional details regarding the training procedures of each of our models.

\subsection{ICoT Training}
\label{appx_sec:icot_details}
Our ICoT training setup follows the practice outlined in the original ICoT paper~\citep{deng2024explicitcotimplicitcot}.

ICoT works by initially presenting explicit chain-of-thought tokens during training, but gradually removing them across numerous ``stages'' (e.g., epochs).
Concretely, the training examples at each epoch may have the following form:
\begin{align*}
(\text{Epoch 1})&\quad a_0a_1a_2a_3 * b_0b_1b_2b_3\%\%\%\ q_0\ldots q_i \ldots q_j \ldots q_k\ldots q_{\tau}\ \#\#\#\#\ c_0\ldots c_7 \\
(\text{Epoch 2})&\quad a_0a_1a_2a_3 * b_0b_1b_2b_3\%\%\%\ q_i \ldots q_j \ldots q_k\ldots q_{\tau}\ \#\#\#\#\ c_0\ldots c_7 \\
(\text{Epoch 3})&\quad a_0a_1a_2a_3 * b_0b_1b_2b_3\%\%\%\ q_j \ldots q_k\ldots q_{\tau}\ \#\#\#\#\ c_0\ldots c_7 \\
\ldots \\
(\text{Epoch N})&\quad a_0a_1a_2a_3 * b_0b_1b_2b_3\%\%\%\ \#\#\#\#\ c_0\ldots c_7
\end{align*}
where $q_i$ are CoT tokens and $\%,\#$ are special delimiters.
These delimiters have no special meaning beyond matching the setup of \cite{deng2024explicitcotimplicitcot}.
Note that after each epoch, the model sees a shorter chain by truncating some tokens, and that by the end, only the operands and final answer remain.

The actual format of our ICoT data is as follows.
Using an example input of $8331 \times 5015$, digits are presented in least-significant digits first, resulting in the following format:
$$1 3 3 8 * 5 1 0 5\vert\vert 5 6 1 4 + 0 1 3 3 8 0 ( 5 6 9 4 2 1 ) + 0 0 0 0 0 0 0 ( 5 6 9 4 2 1 0 ) + 0 0 0 5 5 6 1 \%\%\#\#\#\# 5 6 9 9 7 7 1 4$$

Unlike \cite{deng2024explicitcotimplicitcot}, instead of using a pre-trained 12-layer GPT model, we train a smaller 2-layer, 4-head GPT-based model from scratch, not only to remove any confounding factors from pre-trained knowledge, but also because the 2-layer 4-head architecture is the simplest form in which ICoT succeeds but standard fine-tuning fails.
The training data consists of 80,800 samples, while the validation and test sets each contain 1,000 held out samples.
We train with a learning rate of \texttt{5e-5}, and remove 8 chain-of-thought tokens at every ``stage'' (which in our case is an epoch).
Both training and validation loss converge after 13 epochs, and achieves 100\% accuracy on the test set.

\subsection{Standard Fine-Tuning}
Similar to ICoT, for our standard fine-tuning model, we train a 2-layer, 4-head GPT-based model from scratch, on the same data as ICoT.
We use a learning rate of \texttt{5e-5}, and the input format is $a_0a_1a_2a_3 * b_0b_1b_2b_3 \% \% \#\#\#\# c_0 \ldots c_7$.
All other hyperparameters match those in our ICoT setup.
The model's loss and accuracy plateaus after 13 epochs, it achieves only about 1\% train and validation accuracy, while digit-level accuracy converges at approximately 81\%, and remains the same even after 60 epochs.

Note that scaling the model larger to a 12-layer, 8-head model achieves the same low accuracy at 1\% and digit-level accuracy of 80\%.

%% file: appendix/fourier.tex
\section{Fourier structure in model's weights, activations}
\label{appx_sec:fourier-structure}

Here we provide a deeper dive into the Fourier structure found in the ICoT model's weights and hidden states.
Namely, we analyze the model's embedding weights, final MLP's weights, and last hidden layer:

\begin{enumerate}
    \item Embeddings $\mathcal{E} \in \reals{}^{10 \times d}$
    \item Final layer MLP output weights $W_{out} \in \reals{}^{d_{mlp} \times d}$, given
    $\text{MLP}(\mathbf{x}) = \sigma(W_{in}\mathbf{x})W_{out}$
    \item Final hidden layer $\mathbf{h}^L_t \in \reals{}^{N \times d}$
\end{enumerate}

where $N (=1,000)$ is the size of our validation set.
For the latter two, we first project them onto the model's embedding space: 
\[
\widehat{W_{out}} = (\mathcal{E}W_{out})^\top \in \reals{}^{d_{mlp} \times 10}
\]
\[
\widehat{\mathbf{h}^L} = (\mathcal{E}\mathbf{h}^L)^\top \in \reals{}^{N \times 10}
\]

Each item $X\in\{\mathcal{E}, \widehat{W_{out}}, \widehat{\mathbf{h}^L}\}$ is a collection of row vectors $\mathbf{x}\in\reals{}^{10}$ whose ten entries correspond to digits $n=0,\ldots,9$.

We find that vectors $\mathbf{x}$ are encoded in a low-dimensional trigonometric subspace.

Namely, consider the Fourier expansion
\[
\sum C_n*e^{-2\pi i \tfrac{kn}{10}}, \quad n=0,\ldots,9.
\]

where $C_n (\neq c_k)$ is some constant per $n$.
Following \cite{kantamneni2025language}, we take frequencies $k \in \{0,1,2,5\}$, yielding the real Fourier basis
\[
\Phi(n) =
\left[
\begin{array}{c@{\quad}c@{\quad}c@{\quad}c@{\quad}c@{\quad}c}
\mathbf{1}(n)
& \cos\!\left(2\pi\tfrac{n}{10}\right) & \sin\!\left(2\pi\tfrac{n}{10}\right)
& \cos\!\left(2\pi\tfrac{n}{5}\right)  & \sin\!\left(2\pi\tfrac{n}{5}\right)
& \boldsymbol{p}(n) 
\\[-2pt]
\scriptstyle (k=0) & \scriptstyle (k=1) & \scriptstyle (k=1) &
\scriptstyle (k=2) & \scriptstyle (k=2) & \scriptstyle (k=5)
\end{array}
\right],
\]
where $\mathbf{1}(n)\equiv 1$ (the DC component) and $\boldsymbol{p}(n)\equiv (-1)^n$ (the Nyquist/parity vector). 
The sine terms for $k=0$ and $k=5$ vanish over $n=0,\dots,9$ and are omitted.

Let $F \in 10\times 6$ be a Fourier matrix with rows indexed by $n \in \{0, \ldots, 9\}$ and columns as defined above.

For each row $\mathbf{x}\in\reals{}^{10}$ we fit least squares coefficients
\[
C \;=\; \arg\min_{C\in\reals{}^{6}} \; \|x - FC\|_2^2
\]
and quantify goodness-of-fit using coefficient of determination
\[
R^2(x) \;=\; 1 - \frac{\|x-FC\|_2^2}{\|x - \bar x \|_2^2},
\]
We report the median $R^2$ over the set of rows in each $X$ (i.e., over $d_{\text{mlp}}$ rows for $\widehat{W_{out}}$, $d$ rows for $\mathcal{E}$, and over batch examples for $\mathbf{h}^L$.

In Table~\ref{appx_tab:fourier_fits} we observe strong fits: the per-row medians lie between $0.85$ and $0.99$, indicating that a six-dimensional trigonometric basis over digits captures the vast majority of variance:

We can extend the Fourier bases to include additional terms, for $k=3, 4$, which forms a 8 dimensional basis (excluding sine terms for $k=0, 5$), which leads to perfect $R^2$ fits.

\begin{table}
\centering
\caption{Median $R^2$ of Fourier fits over digits ($n=0\ldots9$).}
\label{appx_tab:fourier_fits}
\begin{tabular}{cccc}
\toprule
Object & Fourier Basis & Rows aggregated & Median $R^2$ \\
\midrule
$\mathcal{E}$           & $k=0, 1, 2, 5$ & $d_{\text{model}}$    & \;\;0.84\;\; \\
MLP $W_{out}$ weights   & $k=0, 1, 2, 5$ & $d_{\text{mlp}}$      & \;\;0.95\;\; \\
$\mathbf{h}^L$          & $k=0, 1, 2, 5$ & batch examples        & \;\;0.99\;\; \\
$\mathcal{E}$           & $k=0, 1, 2, 3, 4, 5$ & $d_{\text{model}}$    & \;\;1\;\; \\
MLP $W_{out}$ weights   & $k=0, 1, 2, 3, 4, 5$ & $d_{\text{mlp}}$      & \;\;1\;\; \\
$\mathbf{h}^L$          & $k=0, 1, 2, 3, 4, 5$ & batch examples        & \;\;1\;\; \\
\bottomrule
\end{tabular}
\end{table}

%% file: appendix/grads_and_losses_12L.tex
\section{Per Token Gradients and Losses: 12-Layer Model}
\label{appx_sec:grads_and_norms_12L}

Even with a larger 12 layer model, the model fails to learn the right long-range dependencies.
Figure~\ref{fig:grad_norms_and_losses_appx} displays the results -- we see the similar patterns as the 2-layer model, in which middle digits never receive the right gradients and loss does not drop.

\input{Figures/grad_norms_and_losses_appx}

%% file: Figures/grad_norms_and_losses_appx.tex
\begin{figure}
\begin{center}
\includegraphics[width=0.99\textwidth]{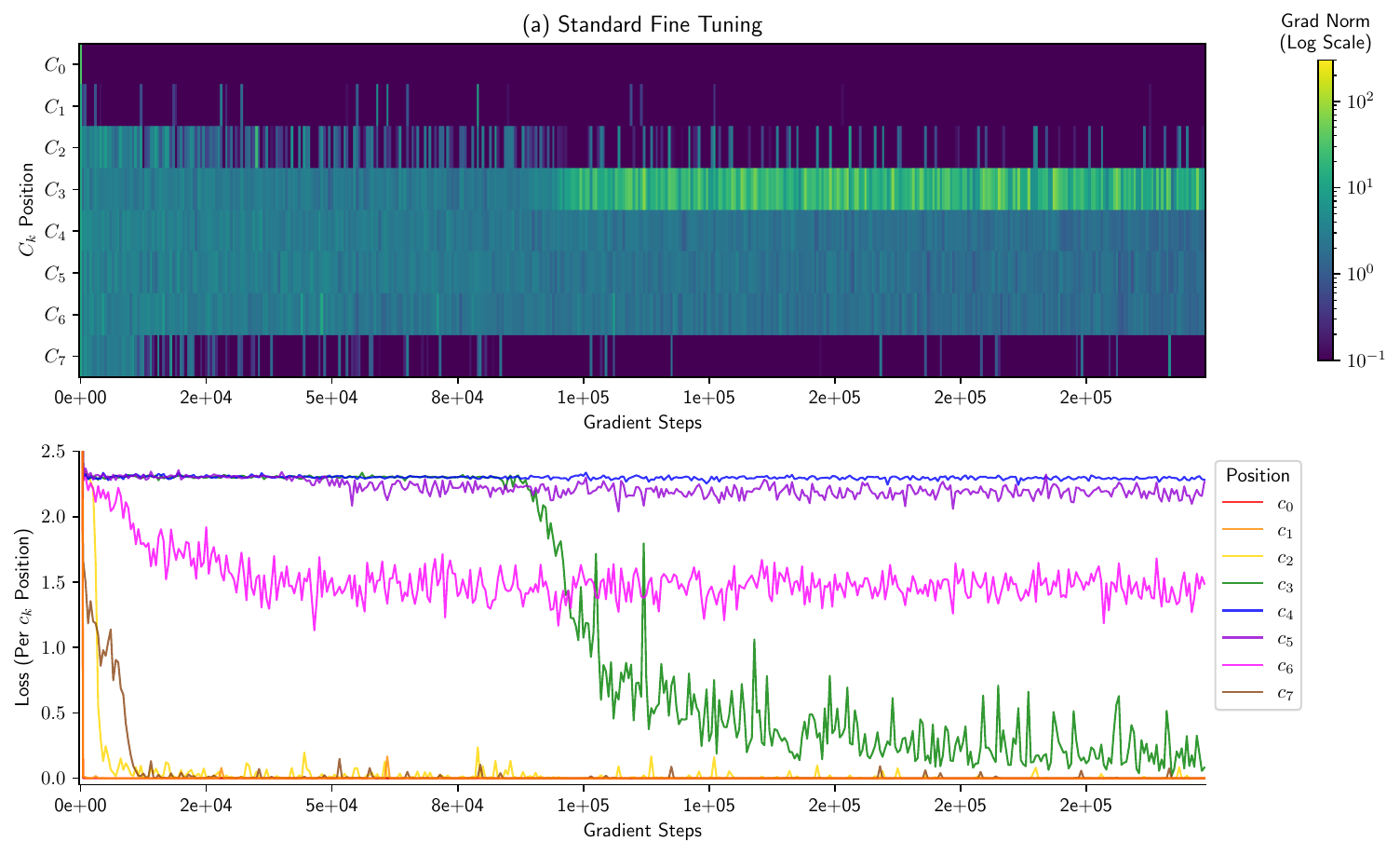}
\end{center}
\vspace{-10pt}
\caption{\label{fig:grad_norms_and_losses_appx}\textbf{Gradients and loss per token for a 12-layer model.}
\vspace{-10pt}
}
\end{figure}

%% file: appendix/attn_tree_dependency.tex
\section{Attention Patterns of All Models}
\label{app_sec:binary_tree}
\input{Figures/attn_tree_appendix} 

In Section~\ref{subsec:binary_trees_via_attention}, we illustrate how a binary tree is constructed for $c_2$ in the ICoT model.
In Figure~\ref{fig:attn_tree_app}, we present the attention patterns for all digits across the three models, from with attention trees can be derived for the ICoT model for each solution token $c_i$.

%% file: Figures/attn_tree_appendix.tex
\begin{figure}[t]
\centering
{\textbf{ICoT}\par}
  \begin{subfigure}[t]{0.9\textwidth}
  \centering
\includegraphics[width=\textwidth]{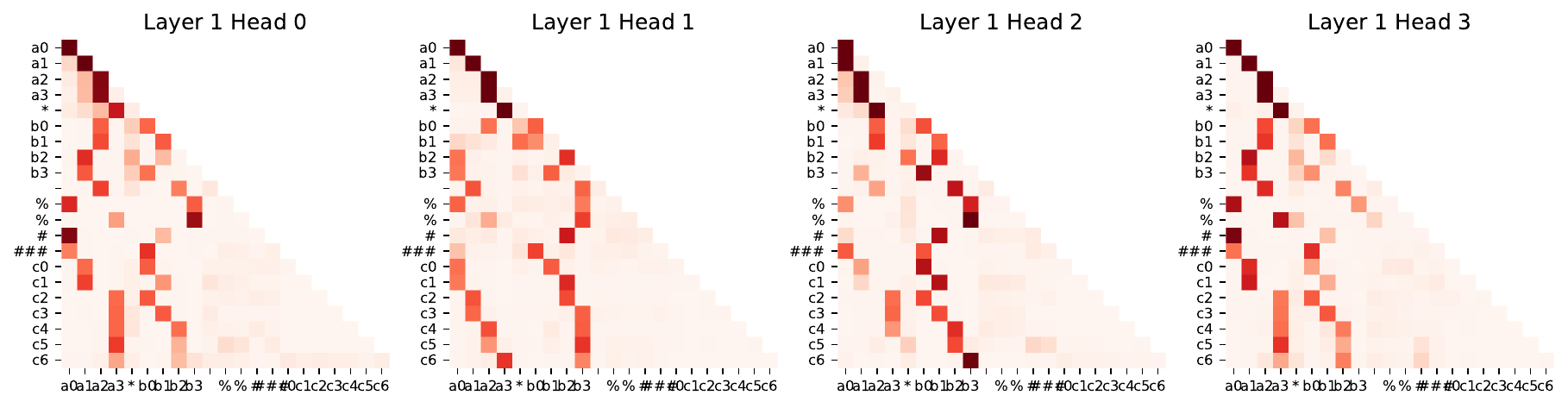}
  \end{subfigure}
  
\vspace{-4pt}
  \begin{subfigure}[t]{0.9\textwidth}
  \centering
    \includegraphics[width=\textwidth]{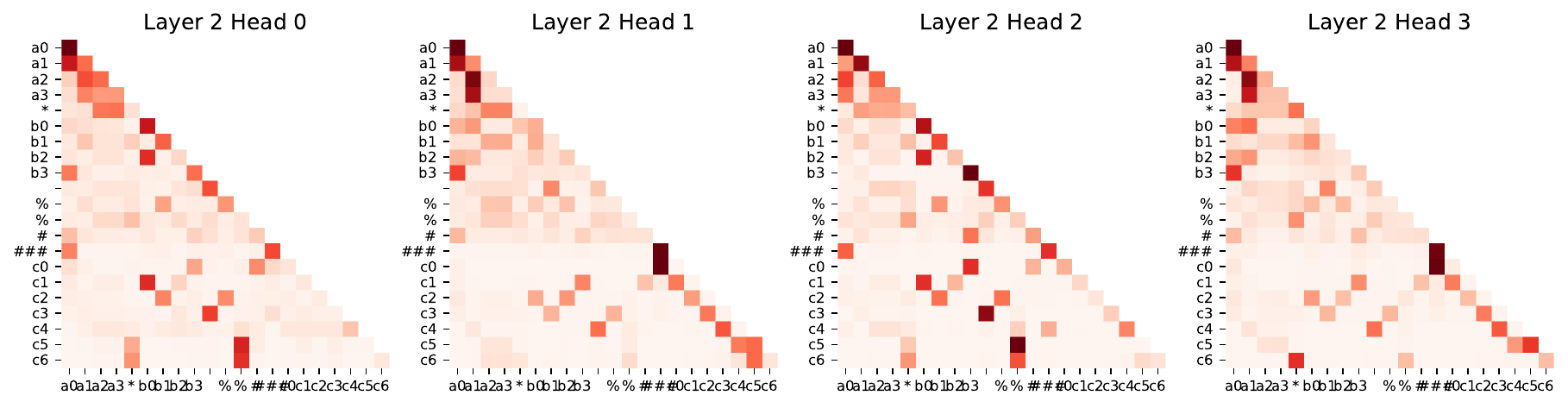}
  \end{subfigure}
  
\vspace{-5pt}
{\textbf{SFT}\par}
  \begin{subfigure}[t]{0.9\textwidth}
  \centering
    \includegraphics[width=\textwidth]{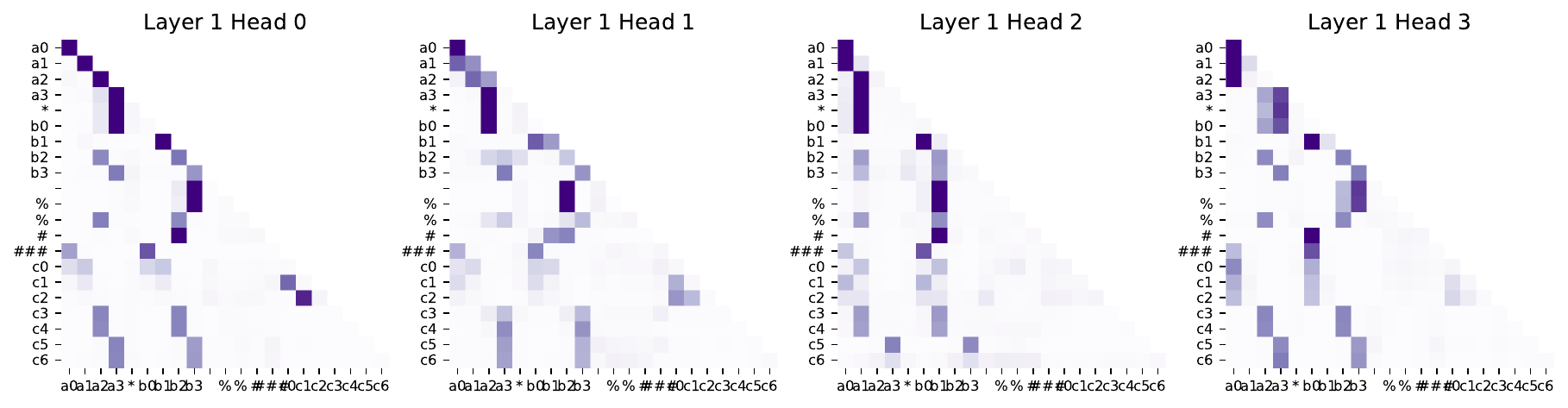}
  \end{subfigure}
  
\vspace{-4pt}  
  \begin{subfigure}[t]{0.9\textwidth}
  \centering
    \includegraphics[width=\textwidth]{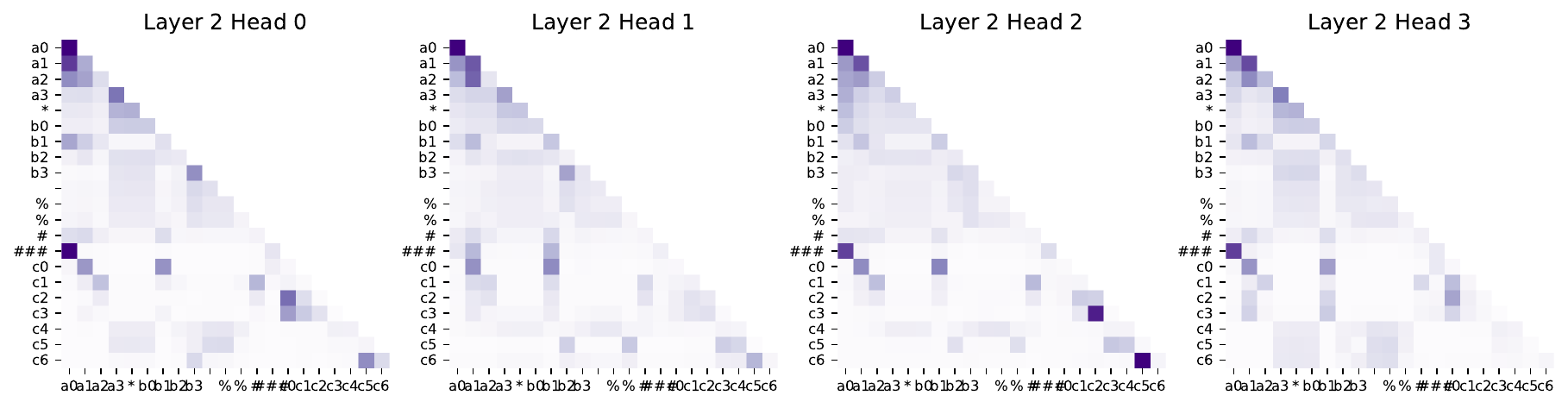}
  \end{subfigure}

\vspace{-5pt}
   {\textbf{Auxiliary Loss Model}\par}
   \begin{subfigure}[t]{0.9\textwidth}
  \centering
    \includegraphics[width=\textwidth]{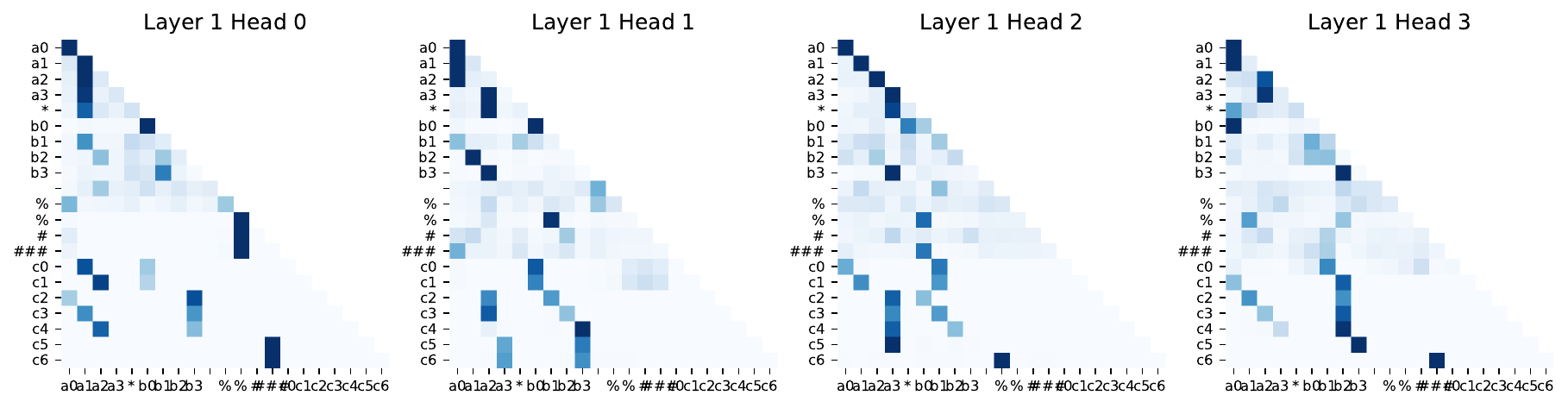}
  \end{subfigure}
  
 \vspace{-4pt} 
  \begin{subfigure}[t]{0.9\textwidth}
  \centering
    \includegraphics[width=\textwidth]{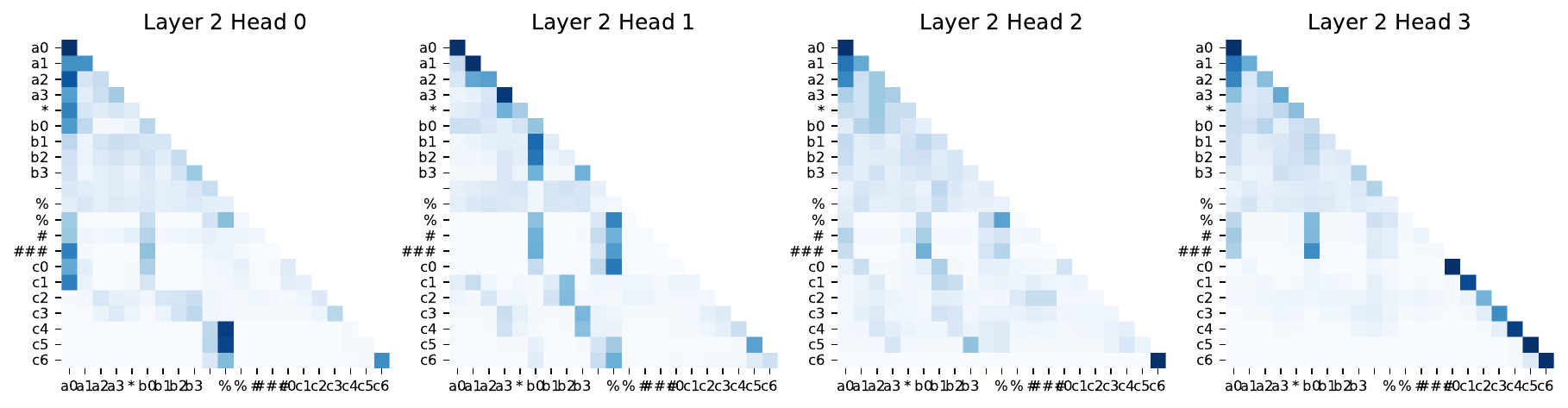}
  \end{subfigure}

\caption{Attention patterns of ICoT, standard fine-tuned model, and the model trained with auxiliary loss }\label{fig:attn_tree_app}
  
\end{figure}

%% file: appendix/llm_usage.tex
\section{LLM Usage}

We used LLMs to proof read our draft and polish our notations.